\documentclass[10pt,twocolumn,letterpaper]{article}
\PassOptionsToPackage{table}{xcolor}
\usepackage[pagenumbers]{iccv}
\usepackage{multirow}
\usepackage{placeins}
\usepackage{microtype}
\definecolor{iccvblue}{rgb}{0.21,0.49,0.74}
\usepackage[breaklinks,colorlinks,allcolors=iccvblue]{hyperref}
\hypersetup{
  pdftitle={Do We Need Learned Targets? Classical Gradients as Supervision for 2D Gaussian Initialization},
  pdfauthor={Kyungjune Baek},
  pdfsubject={arXiv preprint},
  pdfkeywords={Gaussian image representation, Initialization, Analytic supervision, Image gradients, Online target generation}
}

\newcommand{\Tref}[1]{Table~\ref{#1}}

\newcommand{\Sref}[1]{Section~\ref{#1}}

\newcommand{\tabood}{
\begin{table}[tbp]
\centering

\caption{\textbf{Generalization on out-of-distribution data.} Reconstruction quality on Manga109 (drawn art), point-matched at $\sim$199K and optimized for $5K$ iterations. This distribution departs from the DIV2K images on which Instant-GI's targets are optimized. $^{\dagger}$Neural+* is point-unconstrained, listed for reference only.}
\label{tab:ood}

\resizebox{\linewidth}{!}{
\begin{tabular}{lccc}
\toprule
Method & PSNR$\uparrow$ & MS-SSIM$\uparrow$ & \#pts($K$) \\
\midrule
GI         & 28.81 & .992 & 199.4 \\
Instant-GI & 34.89 & .997 & 199.4 \\
\midrule
\rowcolor[gray]{0.95} Analytic (Ours) & 35.87 & .996 & 199.6 \\
\rowcolor[gray]{0.95} Neural (Ours)   & 36.29 & .996 & 199.3 \\
\rowcolor[gray]{0.95} Neural+ (Ours)  & 36.74 & .997 & 198.8 \\
\rowcolor[gray]{0.95} Neural+*$^{\dagger}$ (Ours) & 36.99 & .997 & 221.5 \\
\bottomrule
\end{tabular}
}
\end{table}
}

\newcommand{\tabefficiency}{
\begin{table}[tbp]
\centering

\caption{\textbf{Pipeline efficiency on DIV2K.} A controlled single-resolution sample-count comparison isolating target-generation cost: the official Instant-GI recipe uses three-resolution samples, whereas our reported model uses one. Online generation removes offline target preprocessing; its cost is included in training. Total additionally includes 11s of per-image GaussianImage fitting.}
\label{tab:efficiency}

\resizebox{\linewidth}{!}{
\begin{tabular}{lccc}
\toprule
Method & Pre-proc. & Network & Total \\
\midrule
Instant-GI      & 40,000s & 10,200s & 50,211s \\
Neural (Ours)   & 0s      & 10,200s & 10,211s \\
Analytic (Ours) & 0.3s    & 0s      & 11.3s   \\
\bottomrule
\end{tabular}
}
\end{table}
}

\newcommand{\tabtrajectory}{
\begin{table}[tbp]
\centering

\caption{\textbf{Convergence on Manga109.} PSNR (dB) at increasing optimization budgets (point-matched $\sim$199K). \textbf{Bold} marks the best point-matched method per column; Instant-GI leads at step 100; Neural+ leads from step 250. $^{\dagger}$Point-unconstrained, reference only.}
\label{tab:trajectory}

\resizebox{\linewidth}{!}{
\begin{tabular}{l ccccc}
\toprule
Method & 100 & 250 & 500 & 1000 & 5000 \\
\midrule
Instant-GI & \textbf{28.67} & 30.22 & 31.51 & 32.69 & 34.89 \\
Analytic   & 20.66 & 24.51 & 28.71 & 32.37 & 35.87 \\
Neural     & 24.92 & 29.05 & 31.65 & 33.71 & 36.29 \\
Neural+    & 25.91 & \textbf{30.76} & \textbf{32.98} & \textbf{34.55} & \textbf{36.74} \\
Neural+*$^{\dagger}$ & 29.72 & 32.13 & 33.70 & 34.95 & 36.99 \\
\bottomrule
\end{tabular}
}
\end{table}
}

\newcommand{\tabkvasir}{
\begin{table}[tbp]
\centering

\caption{\textbf{Convergence on Kvasir-SEG.} PSNR (dB) at increasing
optimization budgets ($20K$-point budget, all 1000 images, long side resized to 512).
Kvasir-SEG provides an image-reconstruction evaluation under domain shift. \textbf{Bold} marks the best per column. Under this strong shift, our analytic
prior already leads Instant-GI at \emph{every} budget, and the neural variants lead earlier
still. Random is included as a poor-initialization baseline.}
\label{tab:kvasir}

\resizebox{\linewidth}{!}{
\begin{tabular}{l ccccc}
\toprule
Method & 100 & 250 & 500 & 1000 & 5000 \\
\midrule
Random     & 11.30 & 14.87 & 22.44 & 34.29 & 43.57 \\
Instant-GI & 26.35 & 33.81 & 38.50 & 42.14 & 46.18 \\
\midrule
Analytic   & 33.26 & 38.40 & 41.63 & \textbf{44.13} & \textbf{46.80} \\
Neural     & \textbf{36.50} & \textbf{40.48} & \textbf{42.28} & 43.63 & 45.69 \\
Neural+    & 36.35 & 40.31 & \textbf{42.28} & 43.84 & 45.90 \\
\bottomrule
\end{tabular}
}
\end{table}
}

\newcommand{\figPPM}{
\begin{figure*}[t]
    \centering
    \includegraphics[width=\linewidth]{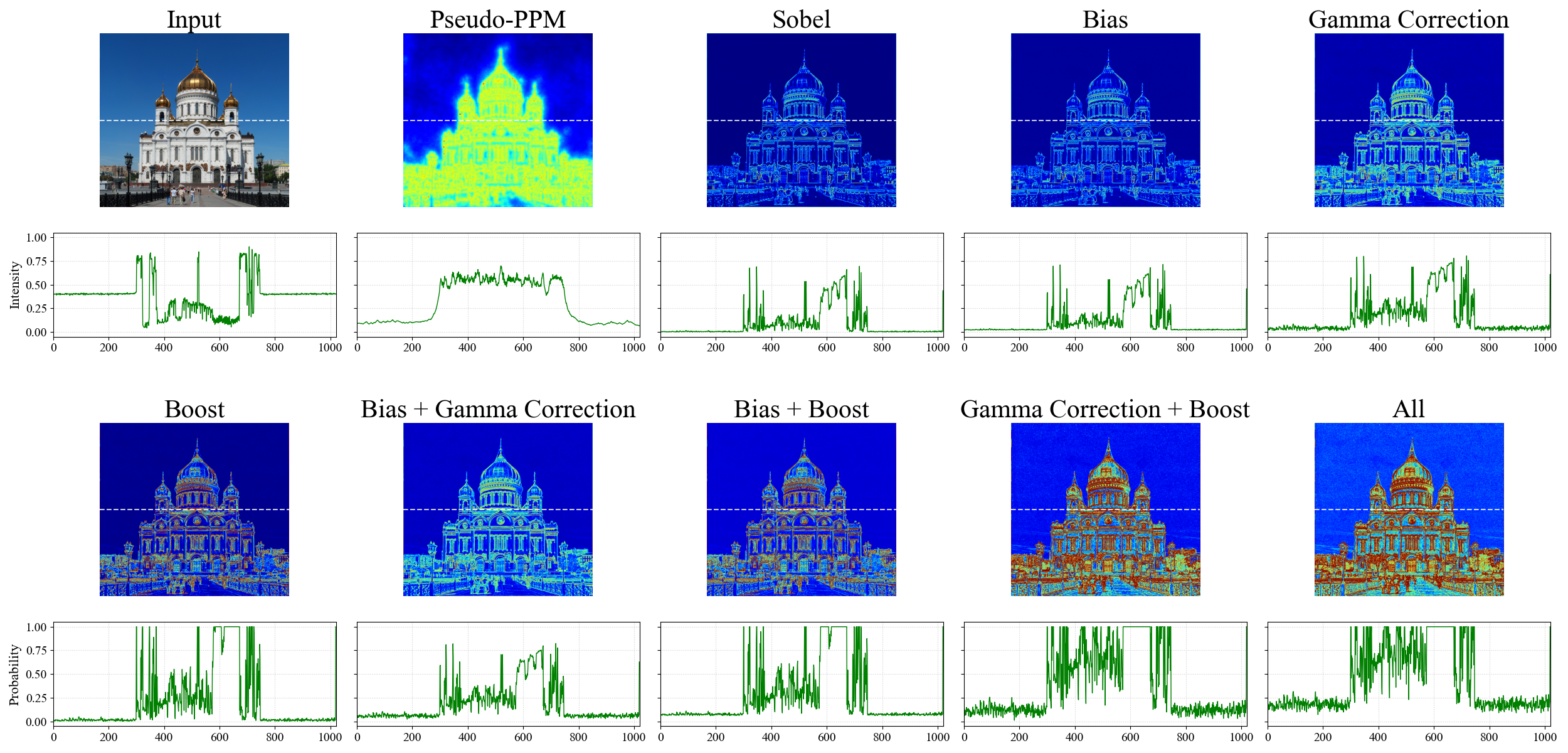}
    
    \caption{\textbf{Visualization of Gradient-Driven Probability Map Generation.} 
We compute an analytic probability map from Sobel gradient magnitudes. For the analytic initialization, we further apply Bias ($\delta$), Gamma Correction ($\gamma$), and Boosting to ensure coverage. The gradient structure serves as a dense supervisory signal for network training, eliminating offline target optimization.}
    \label{fig:compare_ppm}

\end{figure*}
}

\newcommand{\figEvolution}{
\begin{figure*}[t]
    \centering
    \includegraphics[width=\linewidth]{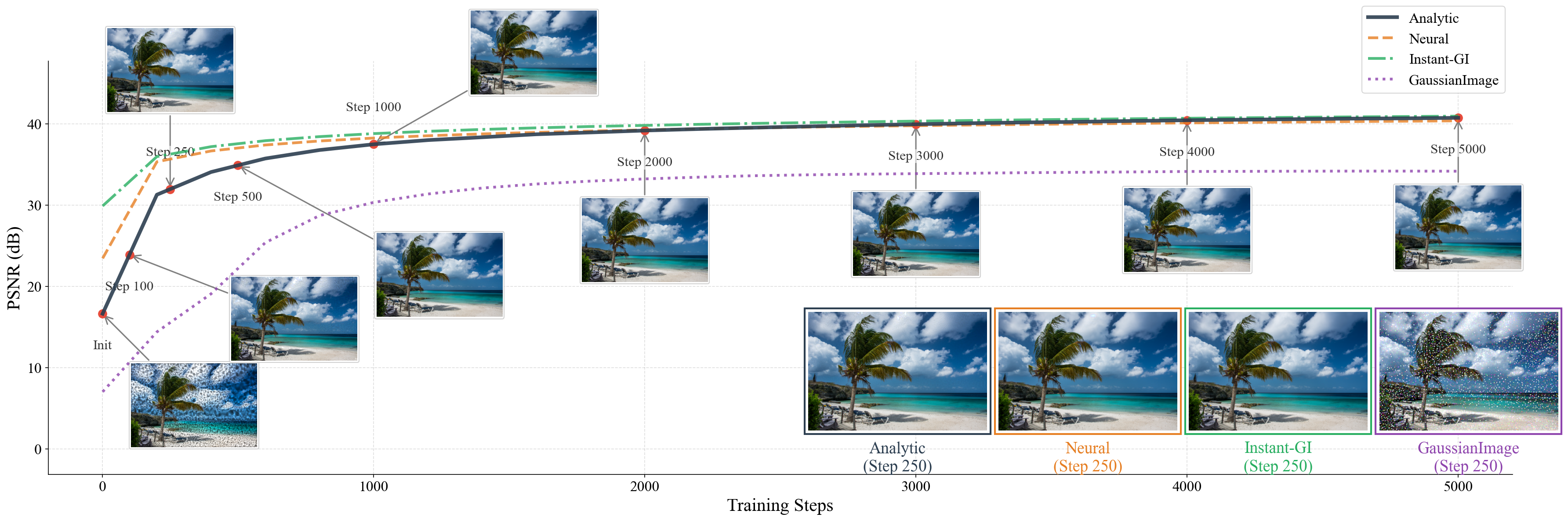}
    
    \caption{\textbf{Training Convergence and Qualitative Progression.} PSNR curves and reconstructions for one DIV2K image. Random initialization (GaussianImage) is included. The plot illustrates fitting behavior on this image.}
    \label{fig:evolution}

\end{figure*}
}

\newcommand{\figOverview}{
\begin{figure*}[t]
    \centering
    \includegraphics[width=\linewidth]{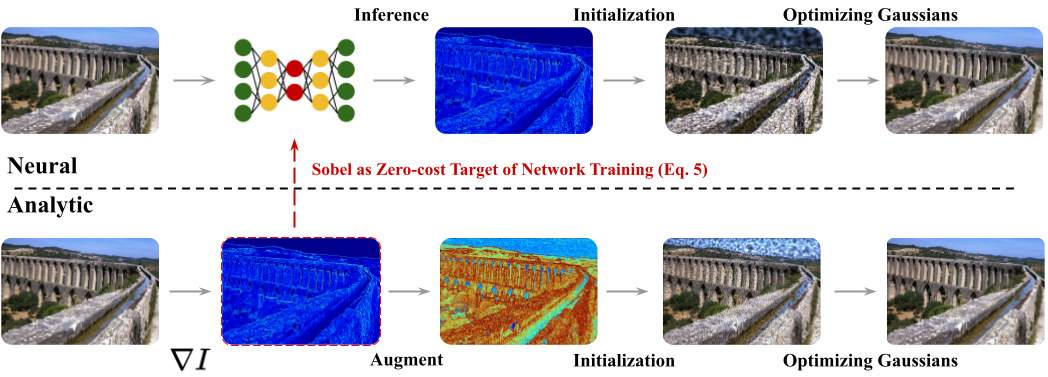}
    
    \caption{\textbf{Overview of the proposed Gaussian initialization framework.} Our framework supports two tracks: (Top) Neural Initialization, which trains a network to predict Gaussian parameters using the raw Sobel map as an online training target (Eq.~\ref{eq:ppm}), and (Bottom) Analytic Initialization, which provides a training-free approach via probability boosting and parallel grid-based sampling. Unlike prior work, both tracks enable online target generation, eliminating the need for expensive pre-optimization.}
    \label{fig:overview}

\end{figure*}
}

\newcommand{\figDither}{
\begin{figure*}[t]
    \centering
    \includegraphics[width=\linewidth]{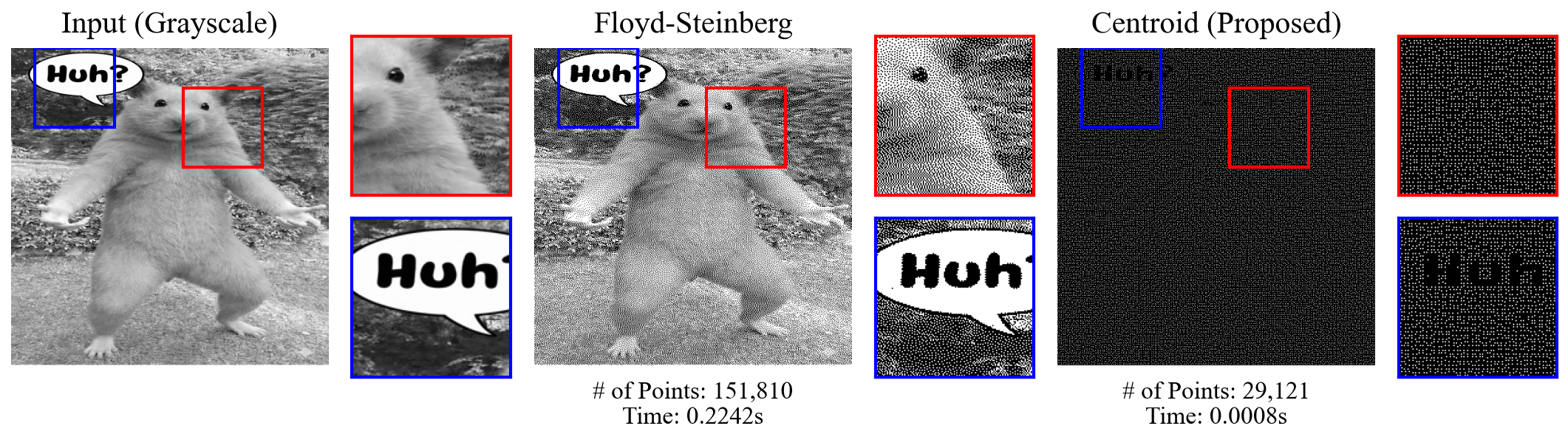}
    
    \caption{\textbf{Comparison of Floyd-Steinberg (FS) dithering and the proposed Grid-based Centroidal Approximation.} The illustrated settings produce different candidate counts (151$K$ for FS and $\sim$30$K$ for the proposed sampler). This image is a qualitative illustration; the matched-budget reconstruction ablation is reported separately.
    }
    \label{fig:dithering}

\end{figure*}
}

\newcommand{\figTargetBudget}{
\begin{figure}[tbp]
    \centering
    \includegraphics[width=0.9\linewidth]{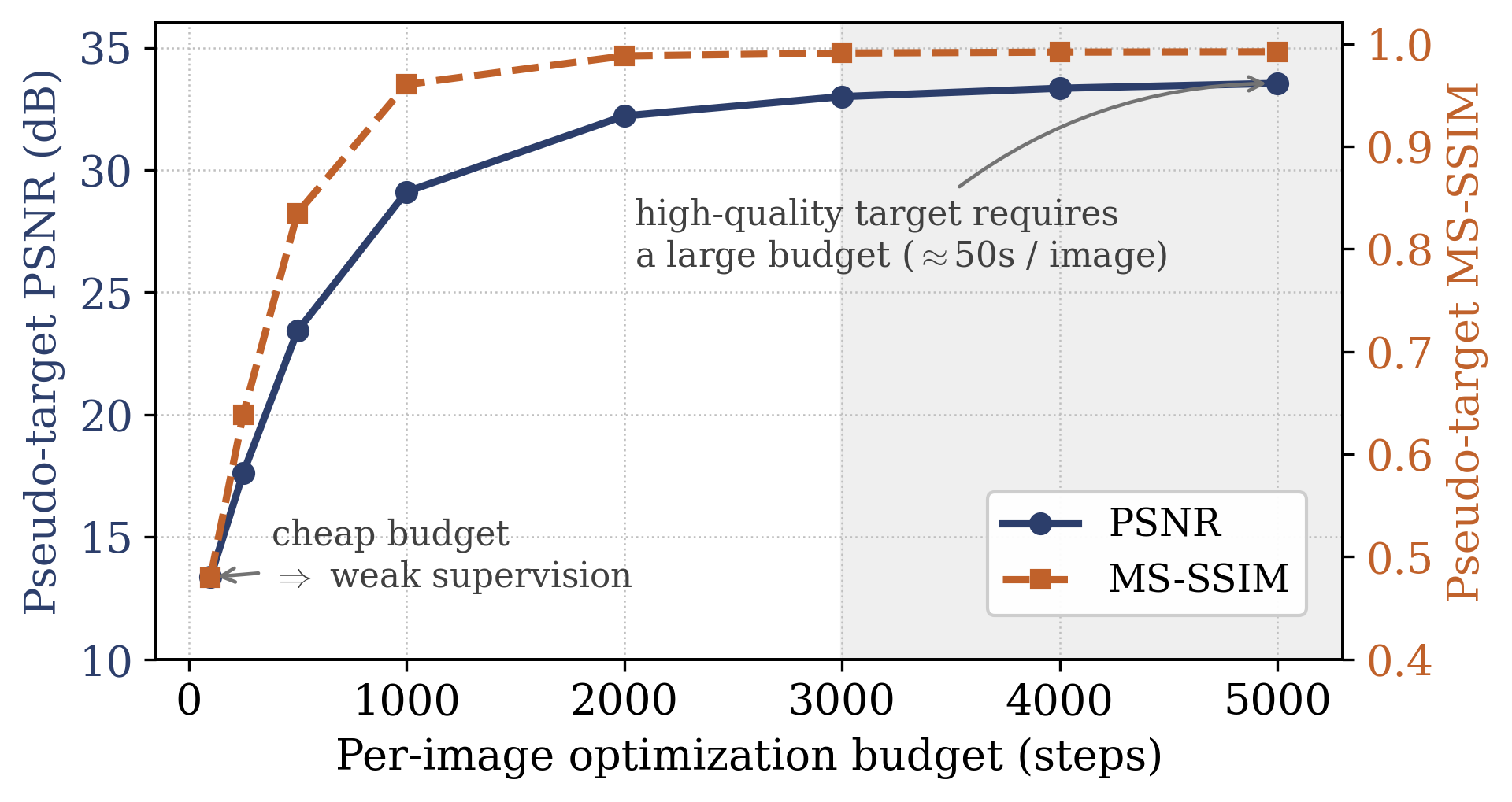}
    
    \caption{\textbf{Cost and supervision quality are entangled.} Quality of the optimized
    Pseudo-PPM target (DIV2K, $50K$-point GaussianImage fitting) as a function of the
    per-image optimization budget. The generating reconstruction improves with the optimization budget ($\approx$50s per image for the reported target-generation setting), illustrating the cost of obtaining optimized supervision. Our analytic target removes this trade-off by construction.}
    \label{fig:target_budget}
    
\end{figure}
}

\newcommand{\tabmain}{
\begin{table*}[t]
\centering
\caption{Reconstruction after $50K$ fitting iterations on Kodak and DIV2K at the requested point budgets $N$. The $43K$ and $77K$ settings correspond to the average Instant-GI counts. PSNR (dB) and MS-SSIM: higher is better. $\dagger$: values reported in~\cite{li2026gaussianimageplus}; other baseline rows are reproduced results. + indicates that the model is trained with 10K images of the LSDIR dataset.}
\label{tab:compare}
\begin{minipage}[t]{0.485\textwidth}
\centering
\small
\setlength{\tabcolsep}{3pt}
\textbf{Kodak}
\par\smallskip\resizebox{\linewidth}{!}{\begin{tabular}{lrrrrrr}
\toprule
 & \multicolumn{2}{c}{$N=5K$} & \multicolumn{2}{c}{$N=10K$} & \multicolumn{2}{c}{$N=43K$} \\
\cmidrule(lr){2-3}\cmidrule(lr){4-5}\cmidrule(lr){6-7}
Method & PSNR & MS-SSIM & PSNR & MS-SSIM & PSNR & MS-SSIM \\
\midrule
GaussianImage$^{\dagger}$ & 29.85 & .962 & 32.48 & .982 & N/A & N/A \\
GaussianImage & 28.53 & .947 & 30.64 & .972 & 38.23 & .996 \\
GI++$^{\dagger}$ & 31.83 & .960 & 35.41 & .983 & N/A & N/A \\
GI++ & 29.75 & .954 & 31.69 & .974 & 35.42 & .986 \\
Instant-GI & 30.28 & .958 & 33.41 & .981 & 42.67 & .997 \\
\midrule
Analytic & 30.32 & .958 & 33.35 & .979 & 42.62 & .997 \\
Neural & 30.61 & .957 & 33.91 & .980 & 43.26 & .997 \\
Neural+ & 30.59 & .956 & 33.94 & .979 & 43.21 & .997 \\
\bottomrule
\end{tabular}}
\end{minipage}\hfill
\begin{minipage}[t]{0.485\textwidth}
\centering
\small
\setlength{\tabcolsep}{3pt}
\textbf{DIV2K}
\par\smallskip\resizebox{\linewidth}{!}{\begin{tabular}{lrrrrrr}
\toprule
 & \multicolumn{2}{c}{$N=10K$} & \multicolumn{2}{c}{$N=50K$} & \multicolumn{2}{c}{$N=77K$} \\
\cmidrule(lr){2-3}\cmidrule(lr){4-5}\cmidrule(lr){6-7}
Method & PSNR & MS-SSIM & PSNR & MS-SSIM & PSNR & MS-SSIM \\
\midrule
GaussianImage$^{\dagger}$ & 26.54 & .910 & 31.45 & .977 & N/A & N/A \\
GaussianImage & 27.27 & .947 & 33.45 & .991 & 35.67 & .994 \\
GI++$^{\dagger}$ & 28.14 & .914 & 33.75 & .978 & N/A & N/A \\
GI++ & 26.26 & .897 & 30.28 & .964 & 31.90 & .974 \\
Instant-GI & 29.95 & .969 & 39.25 & .997 & 42.52 & .998 \\
\midrule
Analytic & 29.63 & .967 & 38.25 & .995 & 41.48 & .998 \\
Neural & 29.87 & .966 & 38.79 & .995 & 41.89 & .998 \\
Neural+ & 29.86 & .965 & 38.83 & .995 & 42.07 & .998 \\
\bottomrule
\end{tabular}}
\end{minipage}

\end{table*}
}

\newcommand{\tabablation}{
\begin{table}[tbp]
\centering
\caption{Analytic-initializer ablations after $5K$ fitting iterations on Kodak and DIV2K. GC denotes gamma correction; Bias is the additive probability offset. PSNR and MS-SSIM: higher is better.}
\label{tab:ablation}
\resizebox{\linewidth}{!}{\begin{tabular}{l @{\hspace{8pt}} cc @{\hspace{10pt}} cc}
        \toprule
        5$K$ iterations& \multicolumn{2}{c}{\textbf{Kodak}} & \multicolumn{2}{c}{\textbf{DIV2K}} \\
        \cmidrule(r){2-3} \cmidrule(l){4-5}
        Method & PSNR$\uparrow$ & MS-SSIM$\uparrow$ & PSNR$\uparrow$ & MS-SSIM$\uparrow$ \\
        \midrule
        - GC \& Bias & 38.94 & .991 & 37.17 & .988 \\
        - $\gamma$ Correction (GC) & 39.58 & .993 & 38.76 & .995 \\
        - Bias & 40.87 & .996 & 39.15 & .995 \\
        Kernel-size ($k$) = 3 & 40.13 & .997 & 38.06 & .997 \\
        Local Variance & 39.34 & .992 & 38.79 & .994 \\
        FS dithering & 40.96	& .996	&39.44	&.997 \\
        \midrule
        Analytic & {40.95} & {.996} & {39.44} & {.997} \\
        \bottomrule
    \end{tabular}}
\end{table}
}

\newcommand{\tabgamma}{
\begin{table}[tbp]
\centering
\caption{Sensitivity to $\gamma$ on Kodak after $5K$ fitting iterations.}
\label{tab:gamma_sweep}
\begin{tabular}{c @{\hspace{15pt}} c @{\hspace{10pt}} c}
        \toprule
        \textbf{Gamma} ($\gamma$) & PSNR$\uparrow$ & MS-SSIM$\uparrow$ \\
        \midrule
        0.1 & 38.54 & .997 \\
        0.2 & 38.59 & .997 \\
        0.3 & 39.41 & .997 \\
        0.4 & 40.42 & .997 \\
        0.5 & 40.92 & .997 \\
        \rowcolor{gray!10} 
        0.6 & \textbf{40.95} & .996 \\
        0.7 & 40.69 & .995 \\
        0.8 & 40.36 & .995 \\
        0.9 & 39.96 & .994 \\
        1.0 & 39.58 & .993 \\
        \bottomrule
    \end{tabular}
\end{table}
}

\title{Do We Need Learned Targets? Classical Gradients as Supervision for 2D Gaussian Initialization}
\author{Kyungjune Baek\\
Sejong University\\
Seoul, Republic of Korea}

\begin{document}
\maketitle
\begin{abstract}
Learning-based Gaussian image initializers can require costly supervision obtained by per-image optimization. We investigate whether the Pseudo-PPM targets used by Instant GaussianImage can be replaced by classical image gradients. Our framework uses normalized Sobel maps in two tracks. A training-free analytic initializer combines probability boosting, parallel grid-based sampling, and geometric processing, producing initial parameters in 0.32 seconds on average. A neural initializer is trained on DIV2K with online Sobel supervision, which removes offline target optimization and allows training to scale to 10,000 LSDIR images without additional preprocessing. Both tracks achieve competitive long-budget reconstruction on Kodak and DIV2K. Under domain shift, the DIV2K-trained neural initializer surpasses Instant-GI on Manga109 from 500 fitting iterations onward (by 1.40 dB at 5K iterations), and training on LSDIR increases this margin to 1.85 dB. On Kvasir-SEG, the training-free analytic initializer exceeds Instant-GI at every reported budget. A controlled pipeline-time comparison yields an estimated $4.9\times$ reduction for DIV2K. These results establish inexpensive analytic targets as a practical alternative to optimization-derived supervision for scalable Gaussian initialization.
\end{abstract}

\section{Introduction}
\label{sec:intro}

Implicit neural representations~\cite{siren,dupont2021coin} model images as continuous functions of spatial coordinates. Their computational cost has motivated explicit representations such as GaussianImage~\cite{zhang2024gaussianimage}, which adapts Gaussian Splatting~\cite{kerbl20233d,huang20242d} to 2D images and enables fast rendering. Fitting these representations from random initialization can require tens of thousands of iterations.

Learned initializers reduce this fitting cost by predicting image-dependent Gaussian parameters. Instant GaussianImage (Instant-GI)~\cite{zeng2025instantgi} predicts a Position Probability Map (PPM) and Gaussian attributes from an input image. Its PPM supervision is generated by optimizing a Gaussian model for each training image and deriving a Pseudo-PPM from the resulting distribution. This preprocessing adds substantial computation and target-storage costs. Other approaches, such as EigenGS~\cite{tai2025eigengs}, learn statistical priors from structurally aligned datasets such as human faces.

\figTargetBudget
\figOverview

The cost of Pseudo-PPM generation depends on the optimization budget used to fit each image. In the $50K$-point DIV2K example in Fig.~\ref{fig:target_budget}, reconstruction PSNR rises from 13.36 to 33.54 dB as fitting increases from 100 to 5000 steps. Reducing this budget changes the reconstruction from which the supervision target is derived. Additional training images and target-generation settings therefore require further preprocessing, limiting the scalability of this approach.

We investigate whether a classical image-gradient operator can provide effective initialization supervision. For 2D images, gradient magnitude provides a direct measure of local intensity variation that can guide Gaussian allocation.

We propose \textbf{Gradient-Driven Gaussian Initialization}, which combines Sobel filtering with parallel grid-based probability sampling. The analytic track applies gamma correction and coverage adjustments to the gradient map, then initializes Gaussians through sampling and geometric processing. It requires no network training and takes 0.32 seconds on average in the reported setting. The neural track uses raw Sobel maps generated online to supervise the initialization network. This removes offline target optimization, so the training set can be expanded from the 800 DIV2K images to a 10,000-image subset of LSDIR~\cite{lsdir} without additional preprocessing.

Our experiments show competitive reconstruction quality on Kodak and DIV2K and improved results under domain shift on Manga109 and Kvasir-SEG. Convergence behavior varies with the dataset and fitting budget; the iteration-wise comparisons report these differences directly. Our contributions are:
\begin{itemize}
    \item We show that classical gradient maps can replace optimization-derived Pseudo-PPMs as supervision for 2D Gaussian initialization.
    \item We introduce parallel grid-based sampling with local centroid placement, reducing the measured discretization time by approximately $27\times$ relative to serial error diffusion while retaining reconstruction quality.
    \item We develop analytic and neural initialization tracks. Online gradient supervision removes target preprocessing and storage, allowing the neural track to train on a 10,000-image LSDIR subset.
\end{itemize}

\section{Related Work}

\subsection{Implicit Neural Representations}
Implicit neural representations (INRs) model images using coordinate-based networks. SIREN~\cite{siren} uses periodic activation functions to represent high-frequency details. COIN~\cite{dupont2021coin} formulates image compression as model fitting, with network weights serving as the compressed representation. COIN++~\cite{dupont2022coin} introduces shared priors and instance-specific modulations to reduce encoding cost. Instant NGP~\cite{muller2022instant} combines neural networks with multiresolution feature grids, and WIRE~\cite{saragadam2023wire} uses wavelet activations to capture fine detail. Evaluating coordinate-based networks during decoding motivates explicit primitives that can be rendered through rasterization.

\figPPM

\subsection{2D Gaussian Splatting for Images}
GaussianImage~\cite{zhang2024gaussianimage} adapts Gaussian Splatting~\cite{kerbl20233d,huang20242d} to image representation using anisotropic 2D Gaussians, achieving decoding speeds exceeding 1000 FPS. GaussianImage++~\cite{li2026gaussianimageplus} improves low-bitrate reconstruction through distortion-driven densification. LIG~\cite{zhu2025large} uses a Level-of-Gaussian strategy with coarse-to-fine residual fitting for large images. SmartSplat~\cite{li2025smartsplat} introduces gradient-color guided sampling for ultra-high-resolution image compression, and Image-GS~\cite{zhang2025image} explores content-adaptive representations for applications including denoising.

These studies, together with earlier adaptive triangulation methods for image approximation~\cite{demaret2006image}, establish the usefulness of content-adaptive primitive allocation. Our work examines the targets used to train an initializer. We use online Sobel supervision in place of optimized Pseudo-PPMs and evaluate both direct analytic initialization and neural initialization trained with these targets.

\subsection{From Learned Priors to Analytic Supervision}
Predictive initializers supply Gaussian parameters before per-image fitting. EigenGS~\cite{tai2025eigengs} learns a PCA-based statistical prior from aligned images. Fast 2DGS~\cite{wang2025fast} and Instant-GI~\cite{zeng2025instantgi} predict image-dependent Gaussian parameters with neural networks. Instant-GI first predicts a PPM, discretizes it, and constructs Gaussian primitives through geometric processing and attribute prediction. Concurrently, AIR~\cite{zeng2026air} trains a feed-forward Gaussian predictor in a self-supervised manner, distilling Gaussian updates obtained by short per-image optimization.

Training Instant-GI requires Pseudo-PPMs derived from per-image Gaussian optimization. The generation and storage of these targets add costs as the training dataset expands. In 3D Gaussian Splatting, Foroutan et al.~\cite{foroutan2024evaluating} similarly examined whether costly SfM point clouds are necessary for initialization. Our framework addresses the corresponding question for 2D initialization: it retains the Instant-GI initializer structure and training objectives while using a directly computed Sobel map as the PPM target.

\figDither
\section{Method}
\label{sec:method}
Our method is built around the use of a normalized Sobel gradient map as the position probability map for 2D Gaussian initialization without any per-image optimization. Figure~\ref{fig:overview} outlines how this map is used in two tracks. The analytic track samples Gaussians directly from a boosted gradient map, requiring neither a trained network nor training data. The neural track uses the raw gradient map as an online target for the PPM head of an Instant-GI-style initializer~\cite{zeng2025instantgi}, replacing its optimized Pseudo-PPM. Both tracks share the same parallel sampling stage, which replaces serial error diffusion, and the same geometric processing, in which the sampled points serve as vertices for triangulation.

\subsection{Gradient-Driven Probability Map Generation}
\label{sec:gradient_ppm}
Given an RGB image $I\in\mathbb{R}^{H\times W\times 3}$, we form grayscale intensities with weights $(0.299,0.587,0.114)$ and apply the standard $3\times3$ Sobel kernels~\cite{sobel1968isotropic}:
\begin{equation}
G_x=I_{\mathrm{gray}}*K_x,\qquad G_y=I_{\mathrm{gray}}*K_y.
\end{equation}
The gradient magnitude is $M=\sqrt{G_x^2+G_y^2}$. Per-image min--max normalization produces $P_{\mathrm{raw}}=(M-\min M)/(\max M-\min M+10^{-8})$. The neural track uses this raw map as its online training target.

\noindent\textbf{Gamma correction and coverage adjustment.}
To retain coverage in weak-gradient regions, the training-free track uses
\begin{equation}
P_{\mathrm{boost}}=\operatorname{clip}\!\left(\beta\left[(P_{\mathrm{raw}}+\epsilon)^{\gamma}+\delta\right],0,1\right),
\label{eq:target_gen}
\end{equation}
with $\gamma=0.6$, $\delta=0.02$, and $\epsilon=10^{-6}$. The analytic initialization entry point sets $\beta=3.0$. The power transform is gamma correction: with $\gamma<1$, it amplifies weak gradient responses. The additive offset provides base coverage, and $\beta$ controls candidate density before budget adjustment. The analytic track uses this adjusted map; the neural track uses $P_{\mathrm{raw}}$.

\subsection{Parallel Grid Sampling and Geometric Initialization}
\label{sec:dithering}
Instant-GI uses Floyd--Steinberg (FS)~\cite{FloydSteinberg} error diffusion to discretize its PPM. We replace that sequential stage with parallel block selection, followed by optional local centroid placement. Figure~\ref{fig:dithering} illustrates the resulting candidate sets.

\tabmain

\noindent\textbf{Block selection.}
We zero-pad the probability map to a multiple of $k$ and partition it into non-overlapping $k\times k$ patches. For a patch $B$, let $E_B=\sum_{(u,v)\in B}P(u,v)$. The implementation selects the patch when $E_B>\tau U_B$, where $U_B\sim\mathcal{U}(0,1)$. Equivalently, the selection probability is
\[
q_B=\min(1,E_B/\tau).
\]
The analytic sampling function uses $P=P_{\mathrm{boost}}$ and $\tau=1$; the neural geometric-processing function uses the predicted PPM and $\tau=0.8$. Thus, the latter threshold increases selection probability by a factor of $1/0.8$ before saturation. Each selected block contributes at most one candidate. Coordinates outside the original image are discarded.

\noindent\textbf{Local centroid placement.}
For a selected patch with positive mass, its candidate coordinate is
\begin{equation}
x_c=\frac{\sum_{(u,v)\in B}uP(u,v)}{E_B},\qquad
y_c=\frac{\sum_{(u,v)\in B}vP(u,v)}{E_B}.
\label{eq:centroid}
\end{equation}
The implementation adds $10^{-6}$ to the denominator for numerical stability. This single centroid computation within fixed grid cells is inspired by centroidal quantization~\cite{lloyd1982least,secord2002weighted}. For $k>1$, the candidate can have fractional pixel coordinates; for $k=1$, it coincides with the selected pixel. We use $k=1$ in the analytic track and $k=3$ in the neural track. Parallel block selection accounts for the speed improvement at $k=1$.

\noindent\textbf{From candidates to Gaussians.}
Following Instant-GI, we add boundary points at an interval based on three times the mean nearest-neighbor distance, remove duplicate vertices, and compute a Delaunay triangulation. An ellipse is fitted to each triangle's three vertices and edge midpoints~\cite{fitzgibbon1999direct}, yielding one Gaussian per triangle. The analytic track uses the fitted ellipse center, scale, and orientation, and samples image colors at its center with a brightness factor of 0.25. Ellipse centers can be sub-pixel even when all sampled vertices are on the pixel grid. In the neural track, a barycentric-coordinate head further refines each Gaussian center inside its triangle, and the remaining heads predict scale, rotation, and color modulation. 

\subsection{Neural Initialization with Online Supervision}
\label{sec:neural_init}
We follow the network architecture and attribute-prediction structure of Instant-GI~\cite{zeng2025instantgi}. Raw Sobel maps are generated in the data loader, allowing additional images to be used directly for training. Neural Init. uses DIV2K, and Neural Init.+ uses the 10,000-image LSDIR subset. 

\noindent\textbf{Optimization objectives.}
We adopt the PPM prediction loss and RGB reconstruction loss from Instant-GI~\cite{zeng2025instantgi}, replacing its Pseudo-PPM target with the online Sobel map:
\begin{equation}
\mathcal{L}_{\mathrm{total}}=\mathcal{L}_{\mathrm{PPM}}+\mathcal{L}_{\mathrm{RGB}}.
\end{equation}
The PPM loss is the focal-weighted MSE used by Instant-GI, based on the weighting principle of~\cite{lin2017focal}:
\begin{equation}
\begin{aligned}
\mathcal{L}_{\mathrm{PPM}}&=\frac{1}{HW}\sum_p e_p^2(1-e^{-e_p})^{f},\\
e_p&=|P_{\mathrm{raw}}(p)-P_{\mathrm{pred}}(p)|,\quad f=2.
\end{aligned}
\label{eq:ppm}
\end{equation}
The reconstruction objective combines weighted mean squared error and a weighted SSIM loss map~\cite{ssim}:
\begin{equation}
\begin{split}
\mathcal{L}_{\mathrm{RGB}}={}&\lambda_{\mathrm{rec}}\left\langle W\odot(I_{\mathrm{GT}}-I_{\mathrm{render}})^2\right\rangle\\
&+(1-\lambda_{\mathrm{rec}})\left\langle W\odot(1-\mathrm{SSIM}(I_{\mathrm{render}},I_{\mathrm{GT}}))\right\rangle,
\end{split}
\label{eq:rgb}
\end{equation}
where $\lambda_{\mathrm{rec}}=0.7$, $W$ is the adaptive reconstruction weight used by Instant-GI, and $\langle\cdot\rangle$ denotes averaging. The PPM objective supervises the predicted probability map, and the reconstruction objective trains the Gaussian attributes.

\noindent\textbf{Scale constraint.}
We replace the soft scale penalty of Instant-GI with a positive bias inside the scale activation:
\begin{equation}
\boldsymbol{s}=(\phi(\boldsymbol{x}_{\mathrm{scale}})+b)\odot\boldsymbol{e}_{\mathrm{size}},\qquad b=0.25,
\end{equation}
where $\phi$ is the inherited positive activation and $\boldsymbol{e}_{\mathrm{size}}$ is the fitted ellipse scale. The bias maintains coverage by imposing a lower bound relative to the local ellipse scale, replacing the soft scale penalty.

\subsection{Point-Budget Adjustment}
\label{sec:budget}
The geometric initializer naturally produces a content-dependent count. The two tracks adjust a requested budget at different stages.

\noindent\textbf{Neural track.}
Given $M$ initialized Gaussians and a requested count $N$, we select $\min(N,M)$ indices without replacement. The selection weights are
\begin{equation}
w_i=0.8\,\mathcal{N}(g_i)+0.2\,\mathcal{N}\!\left(\frac{1}{s_{i,x}s_{i,y}+10^{-6}}\right)+0.02,
\end{equation}
where $g_i$ is the Sobel magnitude sampled at the Gaussian center and $\mathcal{N}$ is min--max normalization across candidates. For this budget-adjustment step, grayscale is the mean of the RGB channels. We normalize the weights to sum to one and sample distinct indices, preserving unique candidate indices. If fewer than $N$ candidates are available, all are retained. The implementation also accepts a retained fraction of the candidate count. We retain the original scales after subsampling, as scale compensation reduced PSNR by 0.2--0.3 dB in our experiments. We note that we apply the same policy to Instant-GI if the adjustment is needed.

\noindent\textbf{Analytic track.}
Before triangulation, the implementation aims for $\lfloor N/2\rfloor$ vertices, sampling without replacement with weights $P_{\mathrm{boost}}(p_i)+0.01$ when there are more candidates. If candidates are insufficient, it retains them. Boundary insertion and triangulation follow this selection, and duplicate vertices are removed before triangulation. The final count depends on boundary insertion and the resulting triangulation. We report the measured Gaussian counts for the near-matched Manga109 comparison.

\section{Experiments}
\subsection{Experimental Setup}

\noindent \textbf{Baselines.}
We compare with GaussianImage, Instant-GI, and GaussianImage++~\cite{zhang2024gaussianimage,zeng2025instantgi,li2026gaussianimageplus}. Instant-GI is the primary baseline because our method addresses its supervision-generation cost. EigenGS~\cite{tai2025eigengs} relies on aligned training images and failed to converge on the diverse, unaligned images in our experiments.

\noindent \textbf{Datasets.}
Following Instant-GI~\cite{zeng2025instantgi}, we train on the DIV2K~\cite{Agustsson_2017_CVPR_Workshops} training set (800 images). To demonstrate scalability, we also train on a subset of 10,000 randomly sampled images from LSDIR~\cite{lsdir}, without precomputing optimization-derived targets for these images. We denote the neural initializer trained on DIV2K as Neural and the one trained on LSDIR as Neural+.
For evaluation, we use Kodak~\cite{kodak} (24 images) and the DIV2K validation set (100 images) as Instant-GI. To assess robustness under domain shift, we further use Manga109~\cite{mtap_matsui_2017} (drawn art) and Kvasir-SEG~\cite{jha2019kvasir} (RGB endoscopy), both distributionally distinct from natural images.

\noindent \textbf{Implementation Details.}
All experiments use a single NVIDIA RTX 5090 GPU. Following Instant-GI~\cite{zeng2025instantgi}, the initialization network is a UNet~\cite{ronneberger2015u} with a ConvNeXt encoder~\cite{liu2022convnet}. After initialization, Gaussians are fitted following the Instant-GI protocol. 

\subsection{Comparative Evaluation}
\label{sec:quantitative}
\noindent \textbf{Evaluation Protocol.}
Following Instant-GI~\cite{zeng2025instantgi}, we evaluate reconstruction fidelity using PSNR and MS-SSIM~\cite{wang2003multiscale}. We report two regimes: long-budget reconstruction after $50K$ fitting iterations, and convergence at 100--5000 fitting iterations. All methods in a comparison use matched Gaussian counts; for Instant-GI, whose count is content-dependent, we apply the same subsampling as our neural track (\Sref{sec:budget}) to reach the requested budget.

\noindent \textbf{Reconstruction Quality.}
\Tref{tab:compare} reports long-budget reconstruction on Kodak and DIV2K at $5K/10K/43K$ and $10K/50K/77K$ points, respectively. The $43K$ and $77K$ budgets match the average point counts of Instant-GI~\cite{zeng2025instantgi}, and the lower budgets follow the GaussianImage++~\cite{li2026gaussianimageplus} protocol.

Instant-GI is trained using DIV2K images and their optimized Pseudo-PPM targets. The analytic initializer reaches 30.32 dB on Kodak ($N{=}5K$, vs. 30.28) and stays within 0.06 dB at higher counts. Trained with our online target, the neural variants improve on Instant-GI on Kodak across every budget (30.61/33.91/43.26 dB vs. Instant-GI's 30.28/33.41/42.67); on DIV2K, Instant-GI exceeds Neural+ by 0.09--0.45 dB.

\tabood

\noindent\textbf{Generalization on Out-of-Distribution Data.}
Manga109 provides an evaluation on drawn art (\Tref{tab:ood}). On Manga109, with nearly matched point counts ($\sim$199K), the training-free Analytic (35.87 dB) surpasses Instant-GI (34.89 dB), and the neural variants widen the gap (Neural 36.29 dB; Neural+ trained on LSDIR 36.74 dB), improving over Instant-GI by up to 1.85 dB; the point-unconstrained Neural+* (36.99 dB) is shown for reference only. On Kvasir-SEG, the analytic prior leads at step 100 (33.26 vs. 26.35 dB). Full Kvasir trajectories appear in Table~\ref{tab:kvasir}. We use all 1000 images, resize the long side to 512 pixels, and request a $20K$-point budget. This RGB endoscopy dataset complements the line-art evaluation; the task remains image reconstruction. Spatial error maps are analyzed below; Section~\ref{sec:initialization_diagnostics} examines the initial point allocation.

\noindent \textbf{Comparison with Densification-based Methods.}
\Tref{tab:compare} also includes GaussianImage++ (GI++)~\cite{li2026gaussianimageplus}, which improves low-bitrate quality through dynamic distortion-driven densification. Our reproduction with the official GI++ repository gave lower scores than the published results, so we list both the paper values ($\dagger$) and our reproduction, and compare primarily against the $\dagger$ values. At low point counts GI++ is strong (Kodak $N{=}5K/10K$: 31.83/35.41 dB, above our 30.61/33.94), reflecting densification's strength when the budget is tight. GI++ reports no high-point results ($N{=}43K/77K$), and in our reproduction its pruning under-utilizes the budget, so quality saturates well below ours (Kodak $N{=}43K$: 43.26 vs. 35.42 dB). These results show the strength of GI++ at low point counts and of the proposed initializer at higher point counts in our evaluation.

\noindent \textbf{Convergence Across Budgets.}
Table~\ref{tab:trajectory} reports Manga109 results under nearly matched point counts. Instant-GI leads at 100 iterations (28.67 dB); Neural+ overtakes it at 250 iterations (30.76 vs.\ 30.22 dB) and leads the point-matched methods thereafter. The analytic initializer improves more slowly on this dataset: it remains below Instant-GI through 1000 iterations and exceeds it at 5000 iterations (35.87 vs.\ 34.89 dB). 

\figEvolution

\tabtrajectory
\tabkvasir

On Kvasir-SEG, the behavior differs. The analytic initializer exceeds Instant-GI at each reported budget, including 100 iterations (33.26 vs.\ 26.35 dB). The neural variants lead at 100--500 iterations, while the analytic variant is best at 1000 and 5000 iterations. At 5000 iterations, the ranking is Analytic, Instant-GI, Neural+, and Neural. Random initialization is included in the same table: its lower early PSNR illustrates the benefit of informed initialization on this dataset. These trajectories support dataset- and budget-dependent convergence gains.

Figure~\ref{fig:evolution} provides a complementary trajectory on one DIV2K image, including random initialization. This example illustrates the per-image fitting process.

\begin{figure*}[t]
    \centering
    \includegraphics[width=0.98\textwidth]{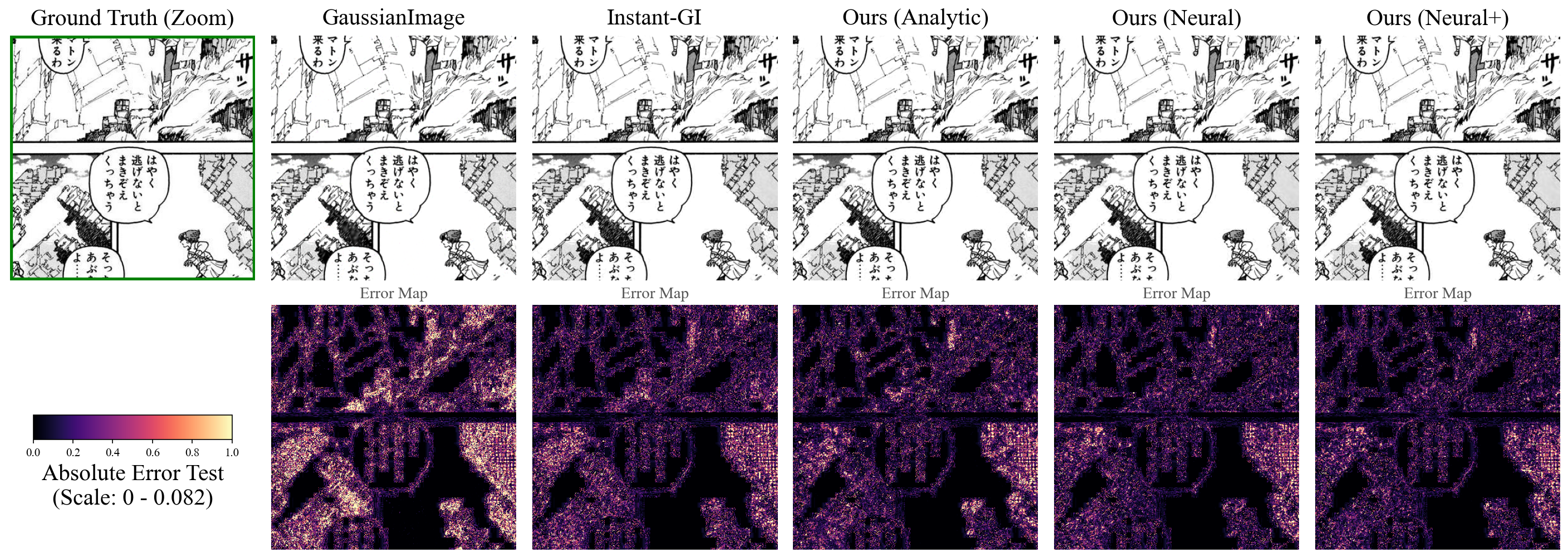}
    \caption{\textbf{Absolute error maps on Manga109.} Our methods (Analytic, Neural, Neural+)
    achieve better fidelity than Instant-GI and GaussianImage. Instant-GI shows error
    concentrations in high-frequency regions (line art, hatching), whereas our methods
    maintain a consistently lower error profile.}
    \label{fig:error}
\end{figure*}

\noindent\textbf{Spatial error analysis.}
\label{sec:spatial_error}
Figure~\ref{fig:error} shows spatial reconstruction errors on a Manga109 example. Instant-GI has concentrated errors in the line-art and hatching regions visible in the crops. The proposed initializers reduce these errors, illustrating the differences in local reconstruction quality.

\subsection{Pipeline Efficiency and Scalability}
\label{sec:efficiency}
\tabefficiency
To demonstrate the efficiency, we compare the total training and inference time against Instant-GI in \Tref{tab:efficiency}. All experiments were conducted on an RTX 5090 GPU. 

\noindent\textbf{Elimination of Pre-processing Bottleneck.} Instant-GI requires approximately 40,000 seconds for pre-processing on DIV2K with 800 training samples. Our online generation removes this offline stage; the Sobel computation is performed during data loading. 

\noindent\textbf{Training Efficiency.} Consequently, the total pipeline time for DIV2K is reduced from 50,211s (Instant-GI) to 10,211s (Ours), achieving a 4.9$\times$ speedup. The speedup mainly comes from elimination of pre-processing (saving ~40$K$ s). For calculating the network training time, we assume that the training set contains a single resolution of the samples. The official checkpoint uses three resolutions; our model uses a single resolution. Even for the large-scale LSDIR dataset (10$K$ images), our total time (52,211s) remains similar to the baseline's time on the much smaller DIV2K dataset. 
Training Instant-GI on the same LSDIR subset would additionally require the large-scale offline pseudo-target generation that our method removes ($\sim$138 GPU-hours for 10K images at $\sim$50s each). This estimate illustrates the additional preprocessing cost avoided by online supervision.

\noindent\textbf{Instant Inference.} For single-image fitting, our Analytic Initialization computes parameters in just 0.32s, offering a high-quality instant initialization of a GaussianImage model.

\noindent \textbf{Storage and I/O Efficiency.}
Online generation also reduces target-storage requirements. Instant-GI must pre-generate
and store Pseudo-PPMs ($\sim$5.27 GB for DIV2K's 800 images), inducing I/O latency during training, whereas online generation requires no persistent target files.

\subsection{Ablation Study \& Analysis}

We validate the effectiveness of our proposed components using the Kodak dataset, as summarized in \Tref{tab:ablation} and \Tref{tab:gamma_sweep}.

\noindent\textbf{Dithering Strategy: Efficiency vs. Quality.} 
As shown in \Tref{tab:ablation}, replacing Floyd-Steinberg (FS) dithering with our Grid-based Centroidal Approximation results in a nearly identical PSNR ($40.96 \rightarrow 40.95$ dB). Discretization time decreases from $0.81s$ for FS to $0.03s$ for parallel block selection and coordinate construction. Combined with the $0.09s$ required for PPM generation, these components avoid per-image target optimization. The complete analytic initializer, including geometric processing, takes 0.32s on average in the reported timing setting.

\tabablation

\noindent\textbf{Impact of Algorithm Components.}
We analyze each module of the Analytic Initialization. Removing gamma correction drops PSNR to 39.58 dB, and removing Bias yields 40.87 dB, primarily affecting flat-region coverage. Removing both, which corresponds to sampling directly from the raw gradient map, reduces PSNR to 38.94 dB on Kodak and 37.17 dB on DIV2K. This is consistent with Instant-GI~\cite{zeng2025instantgi}, which reported that dithering the raw image gradient underperforms its learned PPM because gradient maps concentrate points on edges and leave flat regions sparse. Gamma correction and the coverage offset address this imbalance. Using Sobel gradient magnitude gives higher quality (40.95 dB) than Local Variance (39.34 dB), supporting gradient magnitude as an effective placement proxy in this comparison. Finally, a larger kernel slightly improves initialization speed but degrades quality by limiting spatial granularity; we therefore set $k=1$ for pixel-level candidate selection in the analytic track. Sub-pixel Gaussian centers arise in the subsequent ellipse fitting.

\tabgamma

\noindent\textbf{Sensitivity to Gamma ($\gamma$).} We investigate the impact of the gamma correction term in Eq.~\ref{eq:target_gen}. The results show that performance follows a bell-shaped curve, peaking at $\gamma=0.6$ (40.95 dB). Lower values ($\gamma < 0.3$) flatten the relative contrast between weak and strong responses, while linear mapping ($\gamma=1.0$) does not amplify weak responses; both give lower PSNR in this sweep. The preference for $\gamma<1$ is consistent with asymptotic quantization theory, in which the MSE-optimal point density in two dimensions scales with the square root of the underlying importance density~\cite{gersho1979asymptotically}. Thus, we set $\gamma=0.6$ as the default.

\subsection{Direct Initialization-Quality Diagnostics}
\label{sec:initialization_diagnostics}
To isolate initialization quality from later optimization, \Tref{tab:initialization_diagnostics} measures how
well the initial centers cover structurally important regions, before any reconstruction
loss. On Kvasir-SEG, Analytic attains the highest high-gradient coverage and allocates the smallest fraction of points to low-gradient regions; on Manga109, the neural variants and Instant-GI reach
comparably high coverage, with Analytic showing the most gradient-focused allocation. These metrics use image gradients and therefore favor gradient-driven placement. We use them to characterize the initial allocation of points alongside the reconstruction metrics in Section~\ref{sec:quantitative}.

\begin{table}[t]
\centering
\small
\caption{\textbf{Direct initialization-quality diagnostics.} Metrics on initialized Gaussian
centers \emph{before} reconstruction optimization. HGC: high-gradient coverage (fraction of
top-20\% Sobel-gradient energy covered by centers, $\uparrow$); Emp.: empty high-gradient
cell ratio ($\downarrow$); Low: fraction of points in low-gradient regions.
These metrics characterize the initial point placement. Kvasir on all 1000 images; Manga109 on a 100-image subset (seed 42).}
\label{tab:initialization_diagnostics}
\renewcommand{\baselinestretch}{1}\selectfont
\setlength{\tabcolsep}{3pt}
\resizebox{\linewidth}{!}{\begin{tabular}{l rrr @{\hspace{7pt}} rrr}
\toprule
& \multicolumn{3}{c}{\textbf{Kvasir-SEG}} & \multicolumn{3}{c}{\textbf{Manga109}} \\
\cmidrule(r){2-4} \cmidrule(l){5-7}
Method & HGC$\uparrow$ & Emp.$\downarrow$ & Low & HGC$\uparrow$ & Emp.$\downarrow$ & Low \\
\midrule
Random     & .545 & .0039 & .196 & .593 & .0020 & .283 \\
Instant-GI & .873 & .0002 & .117 & .945 & \textbf{.0000} & .097 \\
\midrule
Analytic   & \textbf{.884} & .0002 & .043 & .929 & .0001 & .020 \\
Neural     & .875 & .0004 & .082 & \textbf{.946} & .0003 & .074 \\
Neural+    & .867 & .0006 & .103 & \textbf{.946} & .0021 & .063 \\
\bottomrule
\end{tabular}}
\end{table}

% \FloatBarrier
\section{Conclusion}
We studied whether 2D Gaussian initialization requires supervision obtained by per-image optimization. A Sobel gradient target supports both a training-free initializer and online supervision of a neural initializer, removing the offline Pseudo-PPM generation stage. Parallel grid-based sampling reduces discretization time by $27\times$, and online target generation allows training on 10,000 LSDIR images without storing optimized targets. The resulting initializers are competitive with Instant-GI on Kodak and DIV2K and outperform it under domain shift, by up to 1.85 dB on Manga109 and at every reported budget on Kvasir-SEG, while the convergence advantage varies with the dataset and fitting budget.

\noindent\textbf{Limitations and Future Work.}
Image gradients are a proxy for useful spatial detail and can also respond to high-frequency noise. Flat regions require a coverage heuristic in the analytic track. Performance varies across datasets and iteration budgets, and the comparisons evaluate the complete initialization pipelines. Future work includes noise-robust targets and initialization of other explicit representations, including video, 3D Gaussian Splatting, and Radiant Foam~\cite{govindarajan2025radiant}.

{\small
\bibliographystyle{ieeenat_fullname}
\bibliography{references}
}
\end{document}